\def\year{2020}\relax
\documentclass[letterpaper]{article} 
\usepackage{aaai20}  
\usepackage{times}  
\usepackage{helvet} 
\usepackage{courier}  
\usepackage[hyphens]{url}  
\usepackage{graphicx} 
\usepackage{graphicx}  
\title{An Analytical Workflow for Clustering Forensic Images: (Student Abstract)}
\author{\Large \textbf{Sara Mousavi}\textsuperscript{\rm 1}, \\ \Large \textbf{Dylan Lee},\textsuperscript{\rm 1}
\Large \textbf{Tatianna Griffin}\textsuperscript{\rm 2}, 
\Large \textbf{Dawnie Steadman},\textsuperscript{\rm 2} 
\Large \textbf{Audris Mockus}\textsuperscript{\rm 1}\\ 
\textsuperscript{\rm 1}Department of Electrical Engineering and Computer Science\\
\textsuperscript{\rm 2}Department of Forensic Anthropology\\
1520 Middle Dr, Knoxville, TN 37996\\
Phone: 8654074882\\ 
Email: mousavi@vols.utk.edu 
}
 
\begin{document}

\maketitle

\begin{abstract}
Large collections of images, if curated, drastically contribute to the quality of research in many domains. Unsupervised clustering is an intuitive, yet effective step towards curating such datasets. In this work, we present a workflow for unsupervisedly clustering a large collection of forensic images. The  workflow utilizes classic clustering on deep feature representation of the images in addition to domain-related data to group them together. Our manual evaluation shows a purity of 89\% for the resulted clusters.

\end{abstract}

\section{Introduction}
Images that show stages of human decomposition, represent high potential value to forensic research and law enforcement. The main sources for such images are forensic anthropology centers and crime scenes. Applications and users benefit from these image collections when labeled with relevant forensic classes, thus improving querying images with the desired content.

Our dataset collected at the University of Tennessee's Anthropology Research Center contains 1 million images collected over 8 years.
Manual labeling is infeasible due to the time and effort required given the sheer number and heterogeneity, different camera angles, body parts and decomposition rates, of the dataset. 
Additionally, creating enough labels for successful supervised learning is also difficult due the scarcity of forensic experts, and the graphic nature of the images.

\begin{figure}
    \centering
    \includegraphics[width=0.3\textwidth, keepaspectratio]{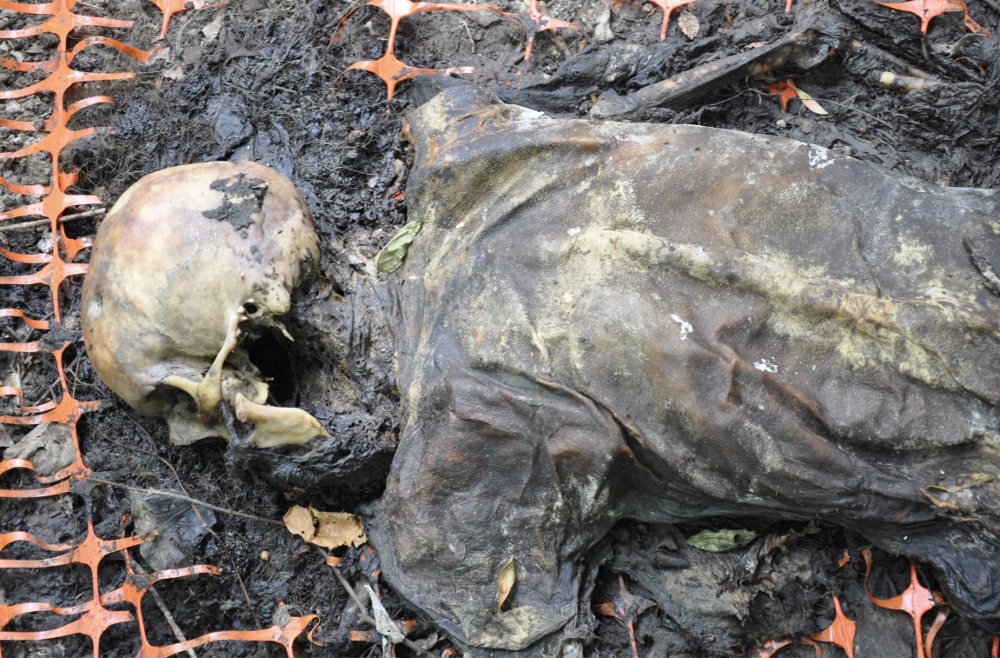}
    \caption{An example image of a decaying body after a month of being exposed to summer weather.} 
    \label{fig:ex}
\end{figure}

Human part detection methods \cite{insafutdinov2017cvpr} do not perform well because of the deformations and decay of the bodies due to environmental factors as well as the natural decay over time. An example image from a dataset containing human decomposition is shown in Figure~\ref{fig:ex}.

In this work, we present an unsupervised analytical workflow, shown in Figure~\ref{fig:arc}, for clustering forensic images. We have found that the key variation in features in such image collections resides along two dimensions: a) different body parts represented in the image and b) different stages of decomposition. Our approach combines information representing domain knowledge about decomposition  with features extracted from the image content to group images along both dimensions. 

First, image features using ResNet50~\cite{he2016deep} pre-trained on ImageNet~\cite{deng2009imagenet} are extracted. However, per our experiments, ResNet does not capture decomposition-related aspects that are absent from ImageNet. We, therefore, incorporate decomposition-related metadata such as temperature, humidity, and wind speed using Accumulated Degree Days (ADD)~\cite{megyesi2005using}. Next, we use t-SNE~\cite{maaten2008visualizing} to get a sense of the number of potential clusters in the data and then cluster the features. Finally we manually evaluate our clustering method using a web interface we designed.

Using our method, we were able to cluster $8507$ images into 15 clusters with an average precision of $89\%$.

\section{Method}
\begin{figure*}
    \centering
    \includegraphics[width=0.9\textwidth, height=4cm]{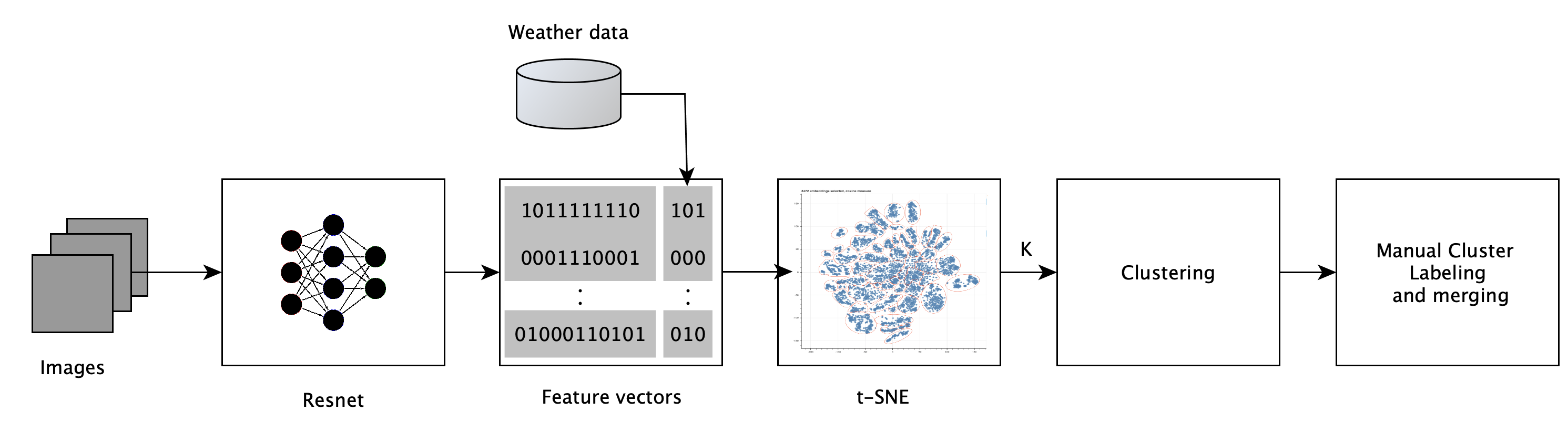}
    \caption{The overall architecture of our workflow is shown. Images are fed into a ResNet model to obtain feature vectors. The features are then combined with weather data. Using t-SNE we find the potential number of clusters in the data and then perform clustering using KMeans. The resulting clusters are then manually labeled and merged. The workflow is implemented using Keras, Python3, HTML, and Javascript. We used a Quadro M6000 GPU for generating the ResNet feature vectors.}
    \label{fig:arc}
\end{figure*}

The developed workflow is shown in Figure \ref{fig:arc}
We exclude the top layers of a pre-trained ResNet50 model since the classes of our collection are completely different. The model is pre-trained on  ImageNet~\cite{deng2009imagenet} 
and produces a $2048$ length feature vector for each image which is then reduced to $256$ via PCA~\cite{wold1987principal}.

Our initial approach consisted of directly clustering the feature vectors afterwards. This resulted in clusters that neither were separated by the body parts nor decomposition stages. 
To improve on this, we extended the feature vectors with external metadata. 
In our dataset, the only metadata available for each photo is an anonymous \textit{Id} of the donor and the date the photograph was taken. We use this information in combination with external 
weather data we obtained for geographic location of the body farm and time of the photos.
Hourly temperature, humidity, and wind speed are used to calculate accumulated degree-days, ADD, for temperature, humidity and wind speed. ADD is commonly used to estimate the postmortem interval~\cite{megyesi2005using} in Forensic Anthropology.

In order to include the weather information in our clustering, we generate a numeric vector, $W$, with the size of $3$. Values in $W$ are ADDs for temperature, humidity and wind speed. In this case, ADD for temperature is calculated as $\frac{T_{d1} + T_{d2}+ \cdots+ T_{dn}}{n}$, where $T_{di}$ is the average temperature for the $ith$ day, and $n$ is the total number of the days since decomposition. ADD for humidity and wind speed are calculated in a similar manner. A new representation for each photo, $P_i$, is created by appending the corresponding $W$ to the current vector representation. 


Most clustering techniques require an estimate of the number of clusters. We visualized the generated $ 256 + 3 = 259$ length vectors for the photos in 2D using t-SNE to find the potential number of clusters in feature-space (included in Figure~\ref{fig:arc}). Based on this plot we chose $50$ clusters and use KMeans clustering technique. 

To evaluate the technique we built a web interface supporting browsing through the images in each cluster. We then selected the miss-clustered images by simply clicking on them. Each selection appends the image name to a text file that can be downloaded at the end of evaluation. Counting the number of the miss-classified images, we calculated the precision of the clustering method.
The web interface also allows labeling the clusters with a meaningful keyword. 
The goal of this cluster-level labeling exercise was to group images of the same body part and order them early to late stages of decomposition based on time.

\section{RESULTS and CONCLUSIONS}
 
In order to test our clustering method, we started with $8507$ photos taken from two donors over $127$ and $122$  photography sessions within 8 months.
Merging the initial 50 clusters from the described data resulted in $12$ clusters.

Our findings show that by adding weather features, 
the clustering precision increased to  $89\%$, from the initial approach
that yielded only 
$64\%$.

In conclusion, we developed an analytical workflow that incorporates external metadata with the image feature representations to cluster a large temporal forensic dataset in an unsupervised manner. 
The resulting clusters not only provide a structured way for the users to navigate a large image collection, but also paves the path for providing data for supervised classification, object detection, and semantic segmentation. 
\section{Acknowledgements}
This work was supported by National Institute of Justice Awards 2016-DN-BX-0179 and 2018-DU-BX-0181.

\bibliographystyle{aaai}
\bibliography{refs}
\end{document}